%% file: main.tex
\documentclass[10pt,twocolumn,letterpaper]{article}

\usepackage[pagenumbers]{cvpr} 
\usepackage{overpic}


\input{setup/macros}

\definecolor{cvprblue}{rgb}{0.21,0.49,0.74}
\usepackage[pagebackref,breaklinks,colorlinks,allcolors=cvprblue]{hyperref}
\usepackage{multirow} 

\title{GausSurf: Geometry-Guided 3D Gaussian Splatting for Surface Reconstruction}

\author{{Jiepeng Wang$^{1}$, Yuan Liu$^{2,3}$, 
Peng Wang$^{1}$,
Cheng Lin$^{1}$,
Junhui Hou$^{4}$,
Xin Li$^{5}$,} \\
{Taku Komura$^{1}$,
Wenping Wang$^{5}$} \\
$^1$The University of Hong Kong, $^2$Hong Kong University of Science and Technology,  \\ $^3$Nanyang Technological University, $^4$City University of Hong Kong, $^5$Texas A\&M University
}

\begin{document}

\input{fig/teaser}

\input{sec/0_abs}   
\input{sec/1_intro}
\input{sec/2_lr}

\input{sec/3_method}
\input{sec/4_exp}

\input{sec/5_conclusion}

{
    \small
    \bibliographystyle{ieeenat_fullname}
    \bibliography{main}
}

\end{document}

%% file: setup/macros.tex
\usepackage{xcolor}
\usepackage{colortbl}
\usepackage{graphicx}
\usepackage{subcaption}
\usepackage{lipsum}
\usepackage{arydshln}

\newcommand{\bSigma}{\boldsymbol{\Sigma}}

\newcommand{\best}{\cellcolor{orange}}
\newcommand{\sbest}{\cellcolor{pink}}
\newcommand{\tbest}{\cellcolor{yellow}}

\usepackage{etoolbox} 
\newrobustcmd{\boxsbest}{
  \begingroup
  \setlength{\fboxsep}{0pt}%
  \colorbox{pink}{\strut \hspace{1em}}%
  \endgroup
}

\newrobustcmd{\boxbest}{
  \begingroup
  \setlength{\fboxsep}{0pt}%
  \colorbox{orange}{\strut \hspace{1em}}%
  \endgroup
}

\newrobustcmd{\boxtbest}{
  \begingroup
  \setlength{\fboxsep}{0pt}%
  \colorbox{yellow}{\strut \hspace{1em}}%
  \endgroup
}

%% file: fig/teaser.tex
\twocolumn[{%
        \renewcommand\twocolumn[1][]{#1}
	\maketitle
        \vspace{-12pt}
	\begin{center}
		\centering
            \begin{overpic}[width=0.95\textwidth]{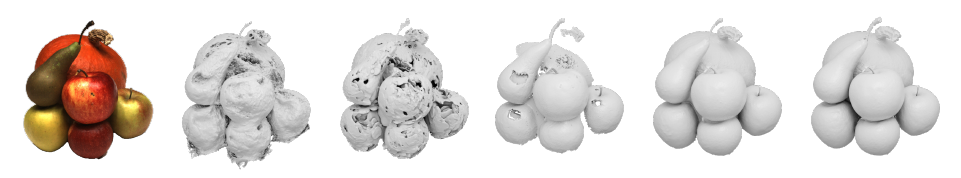}
                \put(3,-2.5){(a) Reference}
                \put(19,-2.5){\begin{tabular}{c}  
                                    (b) 3DGS \\  
                                    11.2m, 1.96
                                    \end{tabular}}
                \put(36,-2.5){\begin{tabular}{c} 
                                (c) SuGaR\\  
                                    $\sim$1h, 1.33  
                                    \end{tabular}}
                \put(52,-2.5){\begin{tabular}{c} 
                                (d) 2DGS\\  
                                    18.8m, 0.80  
                                    \end{tabular}}
                \put(69,-2.5){\begin{tabular}{c} 
                                (e) PGSR\\  
                                    36m, 0.53
                                    \end{tabular}}
                \put(82,-2.5){\begin{tabular}{c} 
                                (f) GausSurf (Ours) \\  
                                    7.2m, 0.52 
                                    \end{tabular}}
          \end{overpic}
          \vspace{12pt}
	   \captionof{figure}{Given a set of posed RGB images (a), our method reconstructs high-quality surfaces (f) with greater efficiency compared to existing Gaussian Splatting-based methods (b-e). In each subcaption (b-f), the first row indicates the reconstruction method, while the second row shows the average reconstruction time and chamfer distance on the DTU dataset, respectively.}
	\label{fig:teaser}
	\end{center}
}]

%% file: sec/0_abs.tex
\begin{abstract}
3D Gaussian Splatting has achieved impressive performance in novel view synthesis with real-time rendering capabilities. However, reconstructing high-quality surfaces with fine details using 3D Gaussians remains a challenging task. 
In this work, we introduce \emph{GausSurf}, a novel approach to high-quality surface reconstruction by employing geometry guidance from multi-view consistency in texture-rich areas and normal priors in texture-less areas of a scene.
We observe that a scene can be mainly divided into two primary regions: 1) texture-rich and 2) texture-less areas.
To enforce multi-view consistency in texture-rich areas, we enhance the reconstruction quality by incorporating a traditional patch-match based Multi-View Stereo (MVS) approach to guide the geometry optimization in an iterative scheme. This scheme allows for mutual reinforcement between the optimization of Gaussians and patch-match refinement,
which significantly improves the reconstruction results and accelerates the training process. 
Meanwhile, for the texture-less areas, we leverage normal priors from a pre-trained normal estimation model to guide optimization.
Extensive experiments on the DTU and Tanks and Temples datasets demonstrate that our method surpasses state-of-the-art methods in terms of reconstruction quality and computation time.
Project page: \url{ https://jiepengwang.github.io/GausSurf/}.

\end{abstract}

%% file: sec/1_intro.tex
\section{Introduction}

Surface reconstruction from multiview images is a long-standing problem in computer graphics and computer vision, which is demanded 
in downstream tasks such as animation, robotics, and AR/VR. Although intensive works have been done on the multiview surface reconstruction task, fast and accurate surface reconstruction still remains an outstanding problem.

Traditional Multi-View Stereo (MVS) methods~\cite{schoenberger2016colmap,shen2013openmvs} are well-established algorithms for multiview surface reconstruction. While MVS methods can achieve high accuracy by pixel-wise matching among different views, they are time-consuming due to their long pipeline including depth estimation, point cloud fusion, surface reconstruction, and texture mapping. Moreover, MVS methods struggle to reconstruct accurate surfaces in areas with low texture due to insufficient features for reliable matching. In recent years, neural rendering methods~\cite{tewari2022advances,tewari2020state}, such as NeuS~\cite{wang2021neus}, VolSDF~\cite{yariv2021volume} and Neuralagelo~\cite{li2023neuralangelo}, provide promising alternatives for multiview surface reconstruction, as these methods enable reconstructing both geometry and appearances in a compact pipeline using a neural shape representation and deliver better reconstruction quality in textureless regions. However, training such neural methods takes extremely long time, typically hours or days, and rendering novel-view images with neural representations is relatively slow.

More recently, 3D Gaussian Splatting (GS) \cite{kerbl3Dgaussians} has emerged as a promising approach to novel view synthesis due to its real-time rendering capability and efficient training time within several minutes. 
However, since 3D GS is chiefly designed for novel-view synthesis instead of surface reconstruction, it does not produce high-quality surface reconstruction. 
SuGaR~\cite{guedon2023sugar} adapts 3D GS for surface reconstruction by regularizng the Gaussians to be more flattened and produces surfaces with noticeable noisy artifacts. 
To improve reconstruction quality, 2DGS \cite{Huang2DGS2024} represents scenes as a set of 2D Gaussian disks and reconstructs high-quality surfaces with surface normal regularizations. PGSR \cite{chen2024pgsr} additionally introduces multi-view photometric regularization into the optimization framework of 3D Gaussian Splatting. However, these methods still face challenges, such as relatively limited reconstruction quality or slow optimization speed.

\input{fig/vis_two_regions_dtu}

To address these challenges, we propose a new 3D Gaussian Surface-based method, GausSurf, for efficient and high-quality multiview surface reconstruction. We observe that natural scenes typically consist of two types of regions: 1) texture-rich and 2) texture-less. For texture-rich areas, we utilize multi-view consistency constraints to guide the optimization process. For texture-less regions, we incorporate normal priors from a pretrained model to provide supplementary supervision signals. By effectively integrating these geometric priors, our method achieves both high-quality and efficient surface reconstruction.

To improve optimization efficiency and accuracy, we enforce the multiview consistency by iteratively incorporating stereo matching into the optimization of Gaussians during the training of GausSurf, in addition to rendering losses of supervision by the input images. 
Specifically, we run the patch-matching algorithm~\cite{barnes2009patchmatch} to refine the depth values and normal maps produced by our GausSurf. This enables us to match multiview images to accurately locate the surface positions during the optimization of the Gaussians.
Subsequently, the enhanced depth and normal will serve as geometry guidance and supervision signals to further guide the optimization of Gaussians. The Gaussian optimization and patch-match refinement are iteratively conducted to enable mutual reinforcement and lead to robust and efficient surface reconstruction. Note that the traditional MVS also utilizes patch-match for depth and normal refinement, but it conducts surface reconstruction in only one single pass, initialized with coarse depth maps and normal maps from sparse SfM points. These initial depth maps and normal maps usually contain significant errors and noise, leading to degraded MVS reconstruction quality. In contrast, our iterative scheme mitigates this issue by seamlessly integrating the optimization of Gaussians and patch-match refinement, resulting in more robust and efficient reconstruction. During the patch-match process, we also adopt an additional geometric verification strategy, where depth or normal values with discrepancies exceeding a robust threshold across multi-views are considered as unreliable and are discarded. These discarded image regions mean that they don't contain sufficient texture for patch-match and cannot produce reliable depth and normal. Thus, we can classify these pixels as texture-less regions, and incorporate additional normal priors as optimization guidance in these areas.

Extensive experiments demonstrated that our method is capable of reconstructing high-quality surfaces on commonly-used datasets, i.e. the DTU~\cite{aanaes2016dtu} dataset and the Tanks and Temples~\cite{knapitsch2017tnt} dataset, and outperforms baseline GS-based surface reconstruction methods in terms of surface quality and reconstruction efficiency. For instance, GausSurf is efficient, costing less than 10 minutes for reconstructing one object with high quality in the DTU dataset. 

We summarize our contributions as follows.
\begin{itemize}
    \item We introduce an efficient framework for high-quality surface reconstruction using 3D Gaussians.
    \item We integrate the traditional MVS algorithm patch matching and normal priors within our framework to enhance reconstruction fidelity and improve computational efficiency.
    \item We demonstrate that our method, GausSurf, has superior speed and quality compared to the state-of-the-art GS-based surface reconstruction methods.
\end{itemize}

We will release the code and data to support future research.

%% file: fig/vis_two_regions_dtu.tex
\begin{figure}
    \centering
    \begin{overpic}[width=\linewidth]{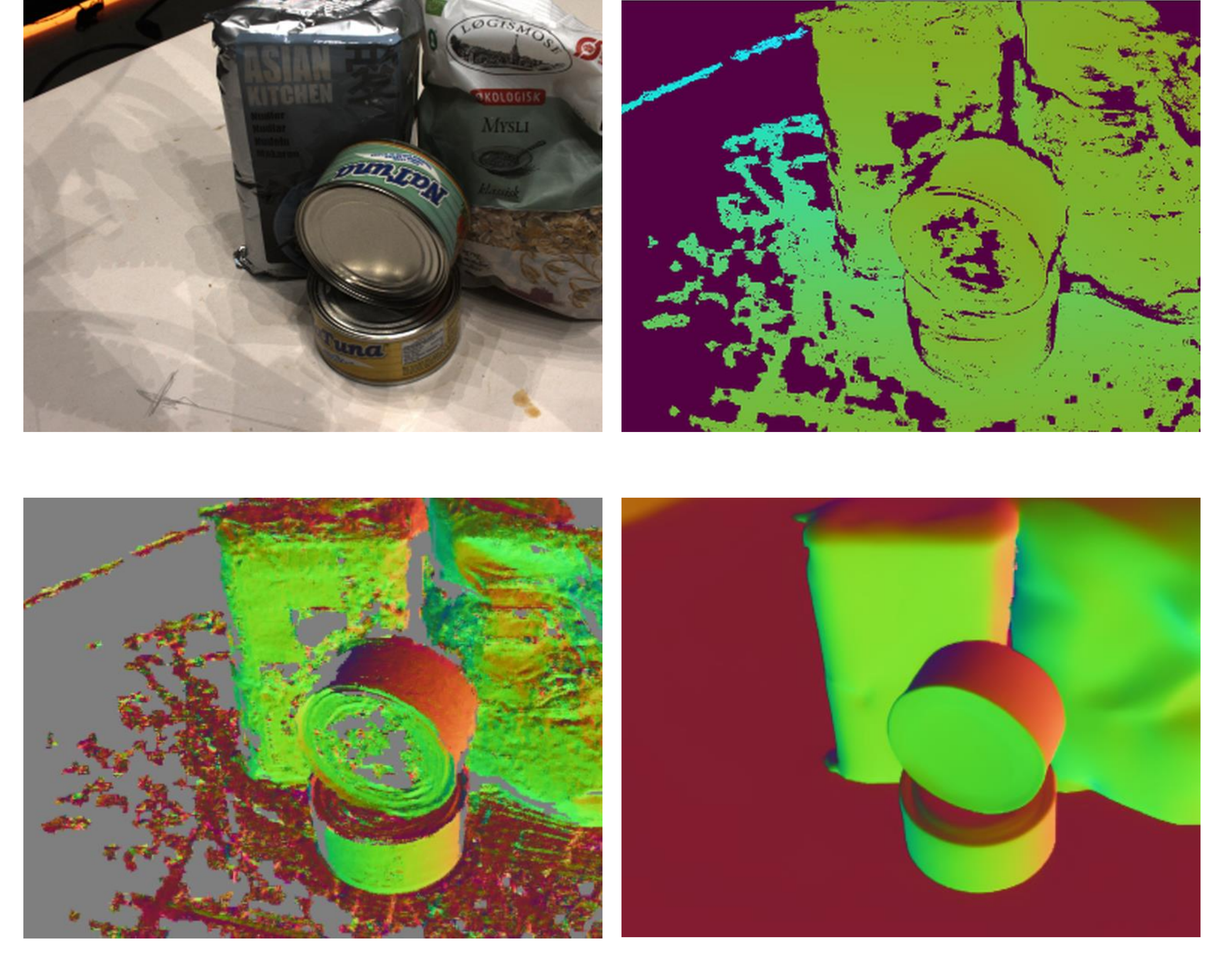}
        \put(15,41){\small (a) Reference}
        \put(65,41){\small (b) PM depth}
        \put(17,-1){\small (c) PM normal}
        \put(63,-1){\small (d) Normal prior}
    \end{overpic}
    \caption{\textbf{Visualization of geometric priors.} (Scene 97 in the DTU dataset \cite{aanaes2016dtu})  (a) Reference image; (b) Refined depth map using patch-match, where the background deep purple color indicates removed unreliable pixels; (d) Refined normal map using patch-match; (d) Estimated normal prior generated by StableNormal \cite{ye2024stablenormal}.}
    \label{fig:vis_two_region_dtu}
\end{figure}
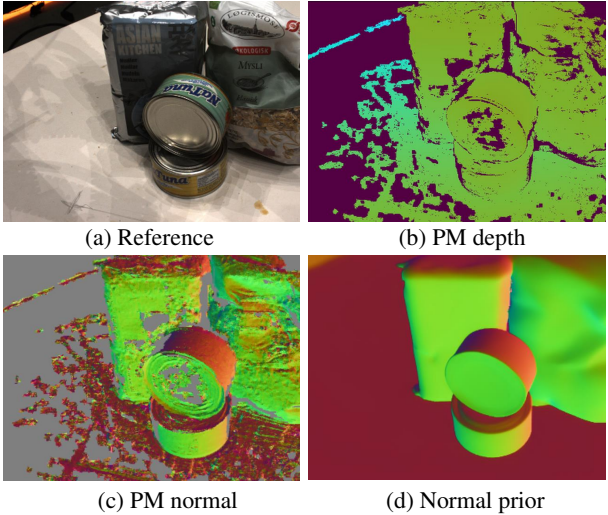

%% file: sec/2_lr.tex
\section{Related works}
In this section, we review works related to multi-view surface reconstruction. Solutions to multi-view reconstruction can be roughly divided into two categories: 1) Multi-view stereo (MVS) methods, which solve for the per-view depth map by maximizing multi-view consistency across views with patch- or feature-level matching and then reconstruct the surface by multi-view fusion;
and 2) differentiable rendering-based methods, which maintain a 3D representation for rendering, allowing rendering errors to be backpropagated so the 3D representation can be optimized. The optimized 3D representation can then be post-processed to obtain the final reconstructed surfaces.

\subsection{Multi-view Stereo}
Classical multi-view stereo typically utilizes patch matching across views for each input image to estimate the depth map. Methods like COLMAP~\cite{schoenberger2016colmap}, OpenMVS~\cite{openMVS}, and PMVS~\cite{furukawa2010accurate} perform well on texture-rich, plain surfaces but degrade in textureless regions and regions with occlusion boundaries.
Learning-based MVS methods, such as MVSNet~\cite{yao2018mvsnet} and its variants~\cite{yao2019recurrent, luo2019pmvsnet, yu2020fast, zhang2023vis}, have addressed the degradation problem in textureless regions. However, they still produce unsatisfactory results at occlusion boundaries and lack multi-view consistency due to the manner of depth prediction for each individual view.

\subsection{Differentiable Rendering}
The recent emergence of NeRF~\cite{mildenhall2020nerf} and its follow-ups~\cite{barron2021mip, barron2022mip, barron2023zip} has demonstrated the power of differentiable rendering for solving the reconstruction and novel view synthesis tasks. For a detailed review, we refer readers to recent surveys~\cite{tewari2022advances, tewari2020state,kato2020differentiable}.

In this paper, we focus on utilizing differentiable rendering for surface geometry reconstruction.

\paragraph{Neural fields as representations}
A neural field~\cite{xie2022neural} typically represents a function mapping a spatial coordinate to values, which are usually approximated by neural networks~\cite{park2019deepsdf, mescheder2019occupancy} or supplemented with feature tables~\cite{liu2020neural, muller2022instant, peng2020convolutional}.
To represent surface geometry, a neural field typically encodes an implicit function where the surface is defined by a level set.
NeRF, though, can inherently encode geometry through volume densities, the surfaces extracted by some isovalue are often of mediocre quality due to the lack of surface constraints.
Alternatively, occupancy fields~\cite{mescheder2019occupancy} or signed distance fields~\cite{park2019deepsdf} are more commonly used for surface reconstruction.
To supervise the neural field with input images, differentiable surface rendering~\cite{yariv2020multiview, niemeyer2020differentiable, lin2022fasthuman, munkberg2022extracting} or differentiable volume rendering~\cite{wang2021neus, yariv2021volume, oechsle2021unisurf} are utilized. However, representing those fields by pure neural networks is often inefficient for both training and inference due to the deep network layers.  

In addition to representations using pure neural networks, several works have utilized hybrid representations with feature tables, such as voxel grids~\cite{cai2023neuda, wu2022voxurf, li2022vox}, triplanes~\cite{wang2023pet}, and hash tables~\cite{cai2023neuda, li2023neuralangelo, wang2023neus2}, to improve training efficiency for fast surface reconstruction.
Although these neural reconstruction methods have proven powerful in surface reconstruction, they produce lower-quality rendering results compared to NeRF-like methods due to the addition of regularizations for surface smoothness. Optimizing these neural surface representations is often very time-consuming with more than 12 hours for a single scene.

\paragraph{Explicit representions}
Compared to representations by neural fields, explicit representations such as point clouds~\cite{kerbl3Dgaussians}, voxels~\cite{fridovich2022plenoxels}, and triangle meshes~\cite{nicolet2021large, worchel2022multi} are more interpretable, efficient, and offer better editability.
More recently, the explicit representation of 3D Gaussian Splatting~\cite{kerbl3Dgaussians}, which comprises a cloud of semi-transparent 3D Gaussian Primitives, has demonstrated remarkable rendering speed, offering more practicality than neural field-based methods, especially on dynamic scenes~\cite{lin2023gaussian, huang2023sc, jiang2024vr} and large-scale scenes~\cite{yu2024sgd, lin2024vastgaussian, yan2024street, cheng2024gaussianpro}.
Unfortunately, while this point cloud-like representation is beneficial for high-quality rendering, it possesses excessive flexibility, making it challenging to derive a high-quality surface from this representation.

The recent work SuGaR~\cite{guedon2023sugar} attempts to improve surface reconstruction quality by introducing additional regularizations, encouraging 3D Gaussian primitives to align with a surface. However, its surface extraction method, which involves externally defining a signed distance function based on the Gaussian primitives, fails to accurately represent the ground truth surface, resulting in bubble-like artifacts.
NeuSG~\cite{chen2023neusg} regularizes the Gaussian primitives to the zero-set of a neural SDF, enabling joint optimization of neural implicit surfaces and 3DGS. However, the reconstruction process is inefficient due to the introduction of neural networks.
GSDF~\cite{yu2024gsdf} and 3DGSR~\cite{lyu20243dgsr} also integrate 3DGS with an extra neural signed distance function.
In contrast, our method does not introduce additional neural networks, with better simplicity and efficiency.
For other works, 2DGS~\cite{Huang2DGS2024} and Gaussian Surfels~\cite{Dai2024GaussianSurfels} utilize 2D Gaussian primitives for better surface alignment, while Gaussian Opacity Fields~\cite{Yu2024GOF} extract the surface by defining an occupancy field derived from the reconstructed 3DGS. 
PGSR \cite{chen2024pgsr} introduces single-view geometric, multi-view photometric and geometric regularizations in Gaussian Splatting's framework to improve reconstruction quality.
We also incorporate multi-view constraints in the optimization but our strategy is different with PGSR and more efficient. Please refer to the supplementary for more discussions about this point and other related works.

%% file: sec/3_method.tex
\section{Method}

Given a set of posed images, our goal is to efficiently reconstruct high-quality surfaces from them while achieving photorealistic novel view synthesis at the same time. 
To achieve this goal, we present a method, called \textit{GausSurf}, that is based on Gaussian Splatting. We regularize 3D Gaussians with multiview stereo (MVS) constraints at texture-rich regions (Sec.~\ref{sec:mvs}) and normal prior guidance at texture-less areas (Sec.~\ref{sec:normal}) to improve reconstruction quality and efficiency. Finally, in Sec.~\ref{sec:losses}, we discuss the loss functions and surface extraction process used in GausSurf.

\subsection{Preliminary}
\label{sec:flatgs}

3D Gaussian Splatting (GS) \cite{kerbl3Dgaussians} represents the scene with a set of 3D Gaussians $\{\mathcal{G}_i\}$. Each Gaussian is parameterized by an opacity $\alpha_i$, its center location $p_i \in \mathbb{R}^3$ and color $c_i \in \mathbb{R}^3$, a rotation $r_i \in \mathbb{R}^4$ in a quaternion form, and a scale vector $s_i \in \mathbb{R}^3$. Thus, the Gaussian distribution in world coordinates is represented by:
\begin{equation}
    \mathcal{G}(x) = e^{-\frac{1}{2} ({x_i-p_i})^T \bSigma^{-1} ({x_i-p_i})}
\end{equation}
where $ \bSigma = \mathbf{R}(r_i)\mathbf{S}(s_i)\mathbf{S}(s_i)^T\mathbf{R}(r_i)^T$ is the covariance matrix consisting of a scaling matrix $\mathbf{S}(s_i)$ and a rotation matrix $\mathbf{R}(r_i)$.

\paragraph{Rendering with 3D Gaussians}
Given a set of 3D Gaussians, RGB images can be rendered via the splatting procedure~\cite{kerbl3Dgaussians} .
We additionally render normal maps and depth maps from 3D Gaussians. 
The normal directions are along the axis with the minimum scaling factor.
To compute the depth value of a camera ray for a specific Gaussian, we adopt the depth value of the intersection point between the camera ray and the plane with the minimum scaling factor of Gaussians \cite{chen2024pgsr}.

\subsection{Patch-match based Geometry Guidance} 
\label{sec:mvs}
Now we explain how to incorporate multiview stereo matching in GausSurf to further improve the reconstruction quality. Specifically, we effectively integrate the patch-match algorithm~\cite{barnes2009patchmatch} in optimizing the Gaussian representation. Such a patch-match algorithm enables our method to consider the consistency of neighboring pixels in a patch while pixel-wise rendering only considers the information on a single pixel. Thus, this patch-match algorithm leads to a more accurate surface reconstruction.

To incorporate the MVS constraints, we propose a \textit{refinement and supervision scheme} in an iterative manner. This approach allows GausSurf to leverage the geometry guidance from MVS to optimize Gaussians and concurrently generate more accurate depth and normal maps, which serve as superior initializations for subsequent MVS refinement. Conversely, from the MVS side, an enhanced starting point contributes to more effective and accurate reconstruction and refinement, which also further provides better geometry guidance for the following Gaussian optimization. By employing an iterative scheme, both representations. i.e., Gaussians and MVS refinement, mutually benefit from each other, resulting in robust reconstruction. Specifically,
we first initialize the Gaussian representation by training with the rendering losses for a predefined number of steps. Then, we render depth maps and normal maps for all training images and feed the rendered depth maps and normal maps to the patch-match algorithm for refinement. After that, these refined depth maps and normal maps will be used to supervise the Gaussian representation for a specified number of steps along with all other losses. After these optimizing steps, we will render again rendering depth maps and normal maps on all images and repeat this refinement and supervision scheme, which is done iteratively until convergence. In the following, we explain this refinement and supervision scheme in more detail. 

\input{fig/method_pm}

\paragraph{Patch-match for refinement} 
Given the rendered depth maps and normal maps from Gaussians, we adopt the patch-match idea~\cite{barnes2009patchmatch, shen2013openmvs} for refinement.
Specifically, for each pixel $r_i$ of every image, we first propagate the depth value $d_i$ and normal direction $\mathbf{n}_i$ of each pixel to its neighboring pixels in a top-to-down and left-to-right order.
Then, for each propagated pixel $r_j$, we evaluate the patch similarities (NCC) $s_j$ and $\hat{s}_j$ with neighboring views on its current depth-normal pair $(d_j, \mathbf{n}_j)$ and the propagated depth-normal pair $(\hat{d}_j, \hat{\mathbf{n}}_j)$, and retain the depth-normal pair with a higher patch similarity.
Note that the propagated $(\hat{d}_j, \hat{\mathbf{n}}_j)$ is augmented with random perturbations before calculating the similarity score.
The same propagation-and-patch-match procedure is then repeated again in reverse order over the maps, i.e., bottom-to-up and right-to-left order, to refine the depth maps and normal maps. 

After propagation and patch matching, additional geometric verification is conducted to check the consistency across the depth and normal maps of different images. If the depth or normal differences between different views are larger than a pre-specified threshold, 
this depth or normal will be regarded as unreliable and removed from the refinement results for this round.
The whole patch-match algorithm refines depth maps according to the patch consistency between neighboring views thus greatly improving the depth quality. Due to the propagation scheme, we only need to evaluate NCC on a small number of depth values, contributing to better efficiency. 

\paragraph{Supervision with refined depth} 
The resulting refined depth maps is then used to supervise the training of the Gaussian representation. Since the geometric verification will discard some unreliable depth values, we only supervise the rendering results from Gaussians at pixels with reliable depth values after geometric verification.
The depth prior loss is defined as the L1 loss between MVS depths and rendered depths from Gaussians:
\begin{equation}
    \ell_d = \sum | d_p - d_i|, 
\end{equation}
where $d_p, d_i$ are the patch-match refined depth and rendered depth, respectively.

\subsection{Normal Prior based Geometry Guidance}\label{sec:normal}
Recent advancements in normal estimation have led to significant improvements in reconstruction quality \cite{yu2022monosdf,wang2022neuris}. 
However, effectively integrating normal priors into GS's framework remains challenging. We observe that normal priors typically provide high-quality estimations in smooth-surface areas by leveraging evident structural information, yet they tend to produce overly smooth estimations in regions with sharp features. This contrasts with patch-match guidance, which excels in these sharp-feature areas, generating high-quality depth maps. With this observation, we find that normal priors and patch-match refinement can be complementary. Patch-match can be applied effectively in texture-rich areas, while normal priors are more suitable for texture-less regions. This raises the question: how can we accurately distinguish between these two types of regions?

As described in Sec. \ref{sec:mvs}, we adopt an additional geometric verification strategy to ensure consistency across the depth and normal maps of different images, where depth or normal values with discrepancies exceeding a robust threshold are considered unreliable and are discarded. We extend this strategy to differentiate pixels in texture-rich and texture-less regions. Pixels passing geometric verification indicate that patch-match can produce accurate depth and normal predictions, and are thus classified as texture-rich. Conversely, pixels failing verification are considered texture-poor, where normal priors are applied as an additional enhancement to achieve high-quality surface reconstruction. Fig. \ref{fig:vis_two_region_tnt} illustrates an example of these distinguished image regions. Through this way, normal priors can be effectively integrated into GausSurf's framework.

\input{fig/vis_two_regions}

\input{table/eval_dtu}

\subsection{Training Losses and Surface Extraction}
\label{sec:losses}

Besides the depth loss in patch-match, we also adopt a color rendering loss $\ell_c$, a depth-normal consistency loss $\ell_{nc}$, and a normal prior loss $\ell_{np}$. The color loss is defined as a combination of L1 error of reconstructed images with a D-SSIM term:
\begin{equation}
    \ell_c = (1-\lambda)\mathcal{L}_1 + \lambda \mathcal{L}_{D-SSIM},
\end{equation}
where $\lambda=0.2$ is used in our experiments. The normal prior loss
is used to regularize the rendered normals at texture-less image regions via normal priors:
\begin{equation}
    \ell_{np} = \sum (1 - n_{p} ^T n_i), 
    \label{eq:dc}
\end{equation}
where $n_{p}$ is the estimated normal prior from StableNormal\cite{ye2024stablenormal}.
The depth-normal consistency loss~\cite{jiang2023gaussianshader,Huang2DGS2024} enforces the consistency between the normals computed from the rendered depth map and the rendered normal maps by
\begin{equation}
    \ell_{nc} = \sum (1 - n_{d} ^T n_i),
    \label{eq:nc}
\end{equation}
where $n_d$ is the normal computed from the rendered depth map while $n_i$ is the rendered normal map.
The total training loss for GausSurf is defined as follows,
\begin{equation}
    \ell = \ell_c + \omega_{nc}\ell_{nc} + \omega_{np}\ell_{np} + \omega_d \ell_d
\end{equation}
where $\omega_{nc}=0.5, \omega_{np}=1.0, \omega_d=1.0$ are predefined constants. 

\paragraph{Surface extraction} After learning the Gaussian representation, we render depth maps from multiview images and fuse these rendered depth maps with a TSDF-fusion algorithm~\cite{izadi2011kinectfusion} to obtain 
the final reconstructed surfaces.

%% file: fig/method_pm.tex
\begin{figure}
  \begin{overpic}[width=\linewidth]{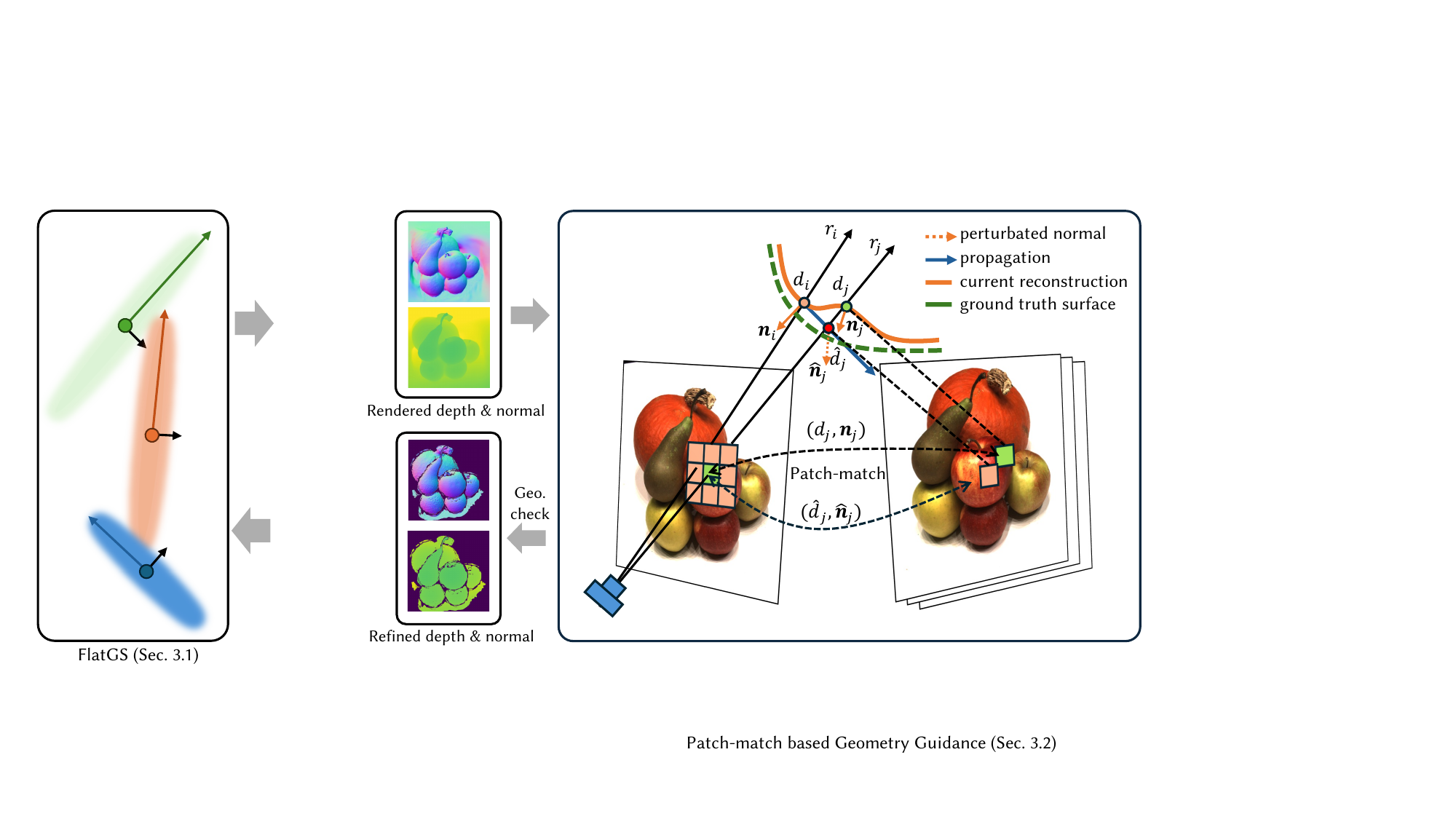}
   \end{overpic}
   
  \caption{\textbf{Patch-match based Geometry Guidance.} Given the rendered depth/normal from Gaussians, we leverage patch-matching to refine the depth and normal for future Gaussian optimization, via propagation, random perturbation and multi-view geometric check.}

  \label{fig:method_pm}
  \vspace{-8pt}
\end{figure}

%% file: fig/vis_two_regions.tex
\begin{figure}
    \centering
    \begin{overpic}[width=\linewidth]{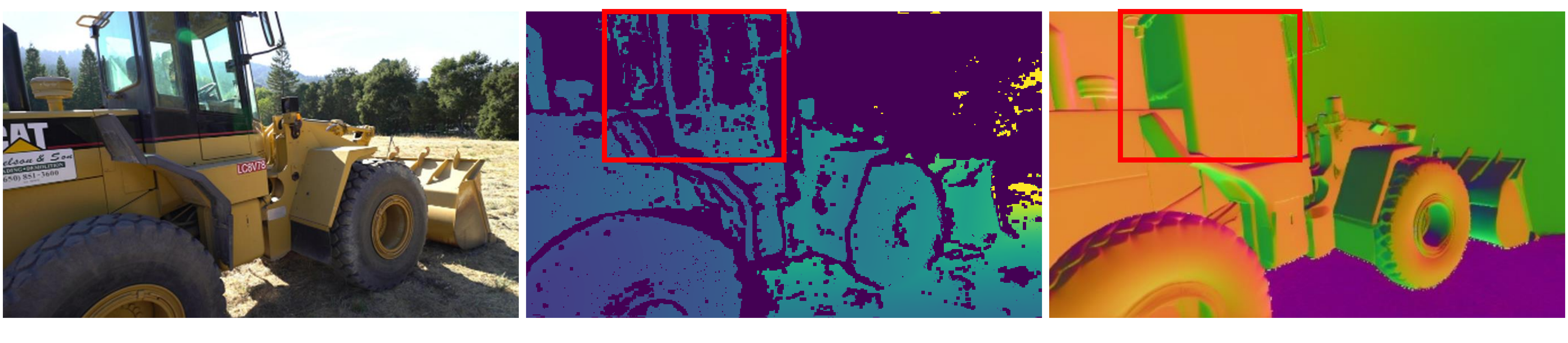}
        \put(7,-2){\small (a) Reference}
        \put(40,-2){\small (b) PM depth}
        \put(71,-2){\small (c) Normal prior}
    \end{overpic}
    \caption{\textbf{Visualization of geometric priors.} (Caterpillar in TnT dataset \cite{knapitsch2017tnt})  (a) Reference image; (b) Refined depth map using patch-match, where the background purple color indicates removed unreliable pixels; (c) Estimated normal prior generated by StableNormal \cite{ye2024stablenormal}.}
    \label{fig:vis_two_region_tnt}
\end{figure}

%% file: table/eval_dtu.tex
\setlength\tabcolsep{0.5em}
\begin{table*}[t]
\centering
\caption{\textbf{Quantitative comparison on the DTU dataset~\cite{aanaes2016dtu}}. We show the Chamfer distance and average optimization time. Our method achieves the highest reconstruction accuracy among other explicit methods. \boxbest, \boxsbest, \boxtbest ~~indicate the best, the second best, and the third best respectively. }
\vspace{-0.2cm}
\resizebox{.98\textwidth}{!}{
\begin{tabular}{@{}llcccccccccccccccclcc}
\hline
 \multicolumn{3}{c}{} & 24 & 37 & 40 & 55 & 63 & 65 & 69 & 83 & 97 & 105 & 106 & 110 & 114 & 118 & 122 & & Mean & Time \\ \cline{4-18} \cline{20-21}
\multirow{4}{*}{\rotatebox[origin=c]{90}{implicit}} & NeRF~\cite{mildenhall2020nerf} & & 1.90 & 1.60 & 1.85 & 0.58 & 2.28 & 1.27 & 1.47 & 1.67 & 2.05 & 1.07 & 0.88 & 2.53 & 1.06 & 1.15 & 0.96 & & 1.49 & $>$ 12h \\
 & VolSDF~\cite{yariv2021volume} & &  1.14 &  1.26 &  0.81 & 0.49 & 1.25 &  0.70 &  0.72 &   1.29 & 1.18 &   0.70 & 0.66 & 1.08 &  0.42 &  0.61 &  0.55 & & 0.86 & $>$12h \\
 & NeuS~\cite{wang2021neus} & &  1.00 & 1.37 & 0.93 &  0.43 &  1.10 &   \tbest 0.65 &    0.57 &  1.48 &  1.09 &  0.83 &  0.52 &  1.20 &  \tbest0.35 &   0.49 &  0.54 & &  0.84 & $>$12h \\
 & NeuralWarp~\cite{darmon2022neuralwarp} & &  0.49 & 0.71 &  0.38  & 0.38  & \tbest 0.79  & 0.81  & 0.82  & 1.20  & 1.06  &  \tbest 0.68  & 0.66  & \tbest0.74 &  0.41  & 0.63  & 0.51  &  & 0.68 & $>$10h \\
 & Neuralangelo~\cite{li2023neuralangelo} & & \tbest 0.37 &  \tbest 0.72 &  \tbest 0.35 & \tbest 0.35 &  0.87 &  \best 0.54 &  \sbest 0.53 &  1.29 &  \tbest 0.97 &  0.73 &  \sbest0.47 &  \tbest0.74 &  \sbest0.32 &  \tbest0.41 &  \tbest0.43 & &  \tbest0.61 & $>$12h \\ 
 \cline{2-2} \cline{4-18} \cline{20-21}
\multirow{6}{*}{\rotatebox[origin=c]{90}{explicit}} 

&  3DGS~\cite{kerbl3Dgaussians}  && 2.14 & 1.53 & 2.08 & 1.68 & 3.49 & 2.21 & 1.43 & 2.07 & 2.22 & 1.75 &  1.79 & 2.55 & 1.53 & 1.52 & 1.50 & & 1.96 &  \sbest{11.2~m} \\
 &  SuGaR~\cite{guedon2023sugar} && 1.47 & 1.33 & 1.13 & 0.61 & 2.25 & 1.71 & 1.15 & 1.63 & 1.62 & 1.07 & 0.79 & 2.45 & 0.98 & 0.88 & 0.79 & & 1.33 & $\sim$1h \\
 & 2DGS~\cite{Huang2DGS2024}     &&  0.48 & 0.91 &  0.39 & 0.39 & 1.01 & 0.83 & 0.81 & 1.36 & 1.27 &  0.76  &  0.70 &  1.40 &    0.40 &   0.76 &   0.52 &&   0.80 &   \tbest18.8~m \\
 & GOF \cite{Yu2024GOF}          && 0.50 & 0.82 & 0.37 & 0.37 & 1.12 & 0.74 & 0.73 &  \tbest1.18 & 1.29 &  \tbest 0.68 & 0.77 &   0.90 & 0.42 & 0.66 &  0.49 &&  0.74 &  2h\\
  & PGSR \cite{chen2024pgsr} && \best0.34 &	\sbest 0.58&	\best0.29&	\best 0.29&	\sbest 0.78&	\sbest 0.58&	\tbest 0.54&	\best 1.01&	\sbest 0.73&	\best0.51&	\tbest 0.49&	\sbest 0.69&	\best 0.31&	\best 0.37&	\sbest0.38&&	\sbest0.53 & 36m \\
 & Ours && \sbest0.35 & \best0.55 & \sbest 0.34 & \sbest 0.34 & \best0.77 & \sbest 0.58 & \best0.51 & \sbest 1.10 & \best 0.69 & \sbest 0.60 & \best 0.43 & \best 0.49 & \sbest0.32 & \sbest0.40 & \best 0.37 && \best 0.52 & \best 7.2 m \\
 \hline
\end{tabular}
}
\label{tab:dtu_result}
\vspace{-0.1cm}
\end{table*}

%% file: sec/4_exp.tex
\section{Experiements}

\subsection{Implementation Details}
All experiments are conducted on a desktop with an i7-13700K CPU and an RTX 3090 GPU.
Specifically, for the DTU dataset, we first train the Gaussian model for 2,000 steps to obtain a rough estimate of the geometry before utilizing patch-match refinement. This preliminary model serves as a solid starting point for the subsequent patch-match-based geometry checks. Patch-match refinement is then performed every 1,000 iterations to refine the depth maps and normal maps to provide geometry guidance. This refinement process continues up to 8,000 steps. 
Following this, we perform another 2,000 iterations to further optimize the Gaussians. This actually leads to 10k optimization steps in total. In the surface extraction with TSDF fusion, we set the voxel size to 0.003 and the truncation threshold to 0.02. Additionally, we utilize the off-the-shelf pretrained method, StableNormal~\cite{ye2024stablenormal}, to obtain normal priors, offering supplementary supervision signals.

\subsection{Evaluation Protocols}
\paragraph{Dataset}
We evaluate GausSurf and the baseline methods on the two commonly used datasets, the DTU~\cite{aanaes2016dtu} and the Tanks and Temples~\cite{knapitsch2017tnt} dataset.
For the DTU dataset, we follow the evaluation protocol of NeuS~\cite{wang2021neus} and 2DGS~\cite{Huang2DGS2024} to evaluate 15 scenes that encompass a wide range of appearances and geometries.
We use the original image resolution to run COLMAP to obtain a sparse point cloud for Gaussian initialization and downsample the images to the resolution of 800$\times$600 to run our algorithm for both surface reconstruction and novel view synthesis. 
Additionally, we also test our method on the Tanks and Temples dataset \cite{knapitsch2017tnt} to verify the effectiveness of our method on large-scale scenes.
For the NVS task, we test our method on the widely used MipNerf360 dataset \cite{barron2021mip}. Following 3DGS, we use one image out of every eight images for evaluation and the remaining seven images for training. 
We report the chamfer distances between the reconstructed surfaces and the ground-truth surfaces as the metrics for surface reconstruction. We also report the NVS quality in terms of PSNR, LPIPS, and SSIM.

\paragraph{Baselines}
We compare our GausSurf with recent representative multi-view surface reconstruction methods. We categorize them into two groups: 1) neural implicit surface reconstruction methods, such as NeRF~\cite{mildenhall2020nerf}, NeuS~\cite{wang2021neus}, VolSDF~\cite{yariv2021volume}, NeuralWarp \cite{darmon2022neuralwarp} and Neuralangelo~\cite{li2023neuralangelo}; and 2) explicit surface reconstruction methods by Gaussian Splatting, including the vanilla 3D Gaussian Splatting~\cite{kerbl3Dgaussians}, and its follow-up work SuGaR~\cite{guedon2023sugar}.
Additionally, we compare our method with other GS-based works for surface reconstruction, namely 2DGS~\cite{Huang2DGS2024}, Gaussian Opacity Field~\cite{Yu2024GOF}, and PGSR~\cite{chen2024pgsr}.

\input{fig/vis_comp_dtu}

\subsection{Results}

\paragraph{Surface reconstruction}
As shown in Tab.~\ref{tab:dtu_result}, our method achieves the best reconstruction results among all explicit Gaussian-based reconstruction methods.
Meanwhile, our method has the lowest reconstruction time among all explicit reconstruction methods with only 7.2 minutes to achieve an accurate reconstruction. The baseline Gaussian-based reconstruction methods commonly take 20 minutes to 1 hour for reconstruction.
In GausSurf, the designed patch-match guidance enables accurate estimation of depth values in texture-rich regions and then propagates them to neighboring regions. Meanwhile, the normal priors can also provide additional supervision signals for the texture-less areas. The geometric guidance provide strong guidance to help the Gaussian representation quickly converge to the correct surfaces and reduce the iterations steps.
When compared to implicit methods such as NeuS~\cite{wang2021neus} and Neuralangelo~\cite{li2023neuralangelo}, our approach achieves competitive reconstruction quality in terms of chamfer distances while requiring significantly less time (7.2 minutes vs. $>$12 hours).
In comparison with NeuralWarp~\cite{darmon2022neuralwarp} which also adopts patch-match in the reconstruction, our patch-match guidance is much more efficient. With additional propagation, random iteration and geometric check strategies for geometry guidance., our model can quickly converge.
Fig. \ref{fig:comp_dtu} presents a qualitative comparison with baseline methods on DTU. Our method can achieve more detailed reconstruction results.

We also quantitatively compare our method with baseline methods on the Tanks and Temples dataset \cite{knapitsch2017tnt}, as shown in Table \ref{tab:tnt}. Our method achieves comparable reconstruction performance to the state-of-the-art method PGSR, while requiring much less computation time.
Compared to GS-based reconstruction method 2DGS~\cite{Huang2DGS2024}, our reconstructed surfaces are much more accurate and smooth with the help of the patch-match and normal prior guidance.  Additionally, we also show the qualitative comparisons with these two methods in Fig. \ref{fig:comp_tnt_mipnerf}. Our method can reconstruct more detailed and accurate surfaces.

\input{table/eval_tnt}

\input{fig/vis_comp_tnt}

\paragraph{Novel view synthesis}
Tab.~\ref{tab:mipnerf360} presents the quantitative comparisons of novel view synthesis quality with state-of-the-art GS-based methods, including 3DGS, 2DGS, and PGSR on the Mip-NeRF360 dataset.
Among all baseline methods and our method, GausSurf showcases comparable image fitting quality and generalization abilities to novel poses.

\subsection{Ablations}
In this section, we ablate the effects of different pipeline modules on reconstruction quality, including the patch-match geometry guidance and normal prior guidance. Table~\ref{tab:ablation_study} and Fig.~\ref{fig:ablation} show the quantitative and qualitative results of the ablation studies respectively. For the two main modules, we ablate their contributions for reconstruction quality respectively:
(a) the patch-match guidance significantly improves the details and accurately locates the correct reconstructed surfaces for GausSurf.
(b) the normal prior guidance can further help improve the reconstruction quality.
As shown in Table \ref{tab:ablation_study}, the reconstruction performance primarily originates from the patch-match. On one hand, this demonstrates the effectiveness of our patch-match strategy; on the other hand, we argue that for the DTU dataset, the scenes predominantly contain texture-rich regions, thus are dominated by the patch-match module. 
We also ablate the effectiveness of the iterative scheme. Please refer to the supplemental for more details.

\input{fig/fig_ablation}

\input{table/ablation_v2}
\input{table/eval_mipnerf}

\subsection{Limitations and Future Work}
Experiments show that our GausSurf method significantly accelerates the training efficiency compared with existing Gaussian-splatting-based methods, especially in object-level reconstruction scenarios. However, GausSurf still requires several minutes per scene optimization, which is not fast enough for real-time applications such as SLAM. In the future, we will explore how to efficiently extend our framework for real-time SLAM reconstruction with streaming video input (e.g., in a feed-forward manner \cite{charatan2024pixelsplat}). 

We observe that GS-based methods critically rely on various optimization strategies to achieve better quality and efficiency. In this regard, we demonstrate that by combining geometric guidance with 3D Gaussian representation, we are able to efficiently achieve high-quality surface reconstruction. It is noticed that, 2DGS \cite{Huang2DGS2024} initialize Gaussians directly as 2D disks for surface reconstruction. It would be interesting to investigate whether appropriate optimization strategies, e.g., iterative patch-match and optimal Gaussian densification strategies could enhance the performance of the 2DGS method. We leave this exploration as future work.

%% file: fig/vis_comp_dtu.tex
\begin{figure*}
  \begin{overpic}[width=\textwidth]{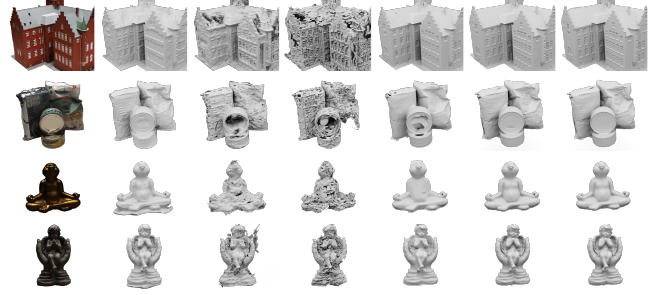}

          \put(2,   -1){\small (a) Reference}
        \put(19, -1){\small (b) NeuS}
        \put(33, -1){\small (c) 3DGS}
        \put(46, -1){\small (d) SuGaR}
        \put(61,-1){\small (e) 2DGS}
        \put(75,-1){\small (f) PGSR}
        \put(90,-1){\small (g) Ours}
   \end{overpic}
   
  \caption{Qualitative comparisons of surface reconstruction on the DTU dataset. Our method can reconstruct more smooth and detailed surface with high efficiency. 
  }
  \label{fig:comp_dtu}
\end{figure*}
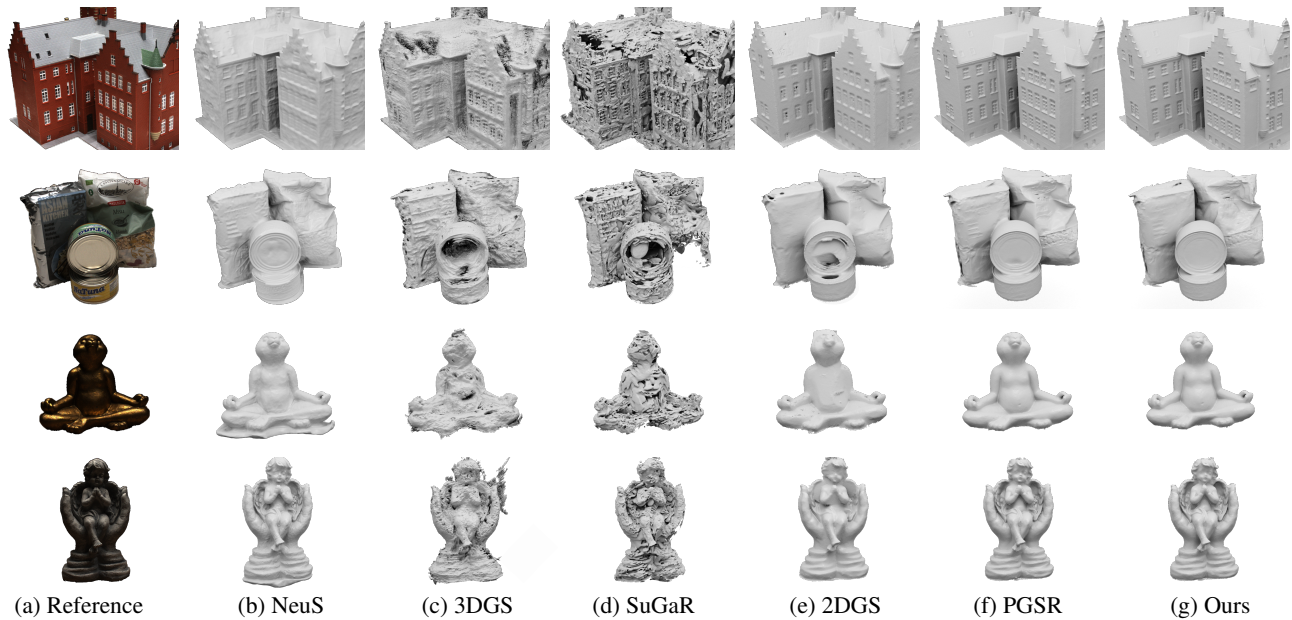

%% file: table/eval_tnt.tex
\begin{table}[t]
\centering
\captionsetup{font={footnotesize}}
\caption{\textbf{Quantitative results of reconstruction results on Tanks and Temples dataset.} Our method achieves comparable reconstruction accuracy (F-score, higher is better) to PGSR with much less optimization time.}
\resizebox{0.98\columnwidth}{!}{
\begin{tabular}{@{}l|ccc|ccccc}
\hline
 & NeuS & Geo-Neus & Neurlangelo &  2D GS & GOF & PGSR & Ours \\ 
 \hline
Barn & 0.29 &  0.33 &   \best 0.70  &  0.36 &  \tbest 0.51 &   \sbest 0.66 & 0.50 \\
Caterpillar &  0.29 & 0.26 &   \tbest0.36 &  0.23 &  \sbest0.41 &  \sbest0.41  & \best0.42 \\
Courthouse &   0.17 & 0.12 &   \sbest0.28 &  0.13 &  \sbest0.28 &  \tbest0.21  & \best0.30 \\
Ignatius &  0.83 & 0.72 &   \best 0.89 & 0.44 & 0.68 &  \sbest0.80 & \tbest0.73\\
Meetingroom & 0.24 & 0.20 &  \sbest 0.32 &   0.16 &  0.28& \tbest 0.29 & \best0.39\\
Truck &  0.45 & 0.45 &   0.48 &  0.26 &   \tbest0.58&  \sbest0.60 & \best 0.65\\ 
\hline
Mean &  \tbest0.38 & 0.35 &   \best0.50 &  0.30 &  \sbest0.46&  \best0.50 & \best0.50\\
Time & \texttt{>}24h & \texttt{>}24h & \texttt{>}128h  &  \best 34.2m &  2h& \tbest 1.2h & \sbest 36.2m \\ 
\hline
\end{tabular}
}
\label{tab:tnt}
\end{table}

%% file: fig/vis_comp_tnt.tex
\begin{figure*}[t]
   \begin{overpic}[width=\textwidth]{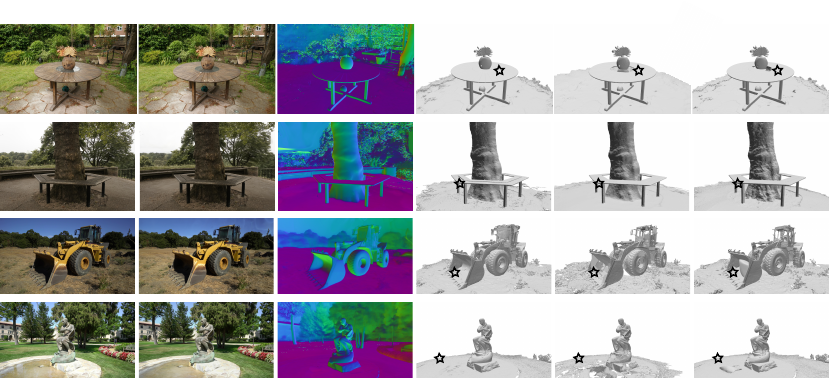}
        \put(3,   -3){(a) Reference}
        \put(18.5, -3){(b) Ours (color)}
        \put(34.5, -3){(c) Ours (normal)}
        \put(55,-3){(d) Ours}
        \put(71,-3){(e) 2DGS}
        \put(88,-3){(f) PGSR}
   \end{overpic}
   \vspace{2pt}
  \caption{Qualitative comparisons on the MipNerf360 dataset (the first two rows) and TnT dataset (the last two rows).  Our method can reconstruct more accurate geometry surface with fine details (See the areas marked with pentagrams).}

  \label{fig:comp_tnt_mipnerf}
\end{figure*}

%% file: fig/fig_ablation.tex
\begin{figure}[!h]
    \centering
   \begin{overpic}[width=0.9\linewidth]{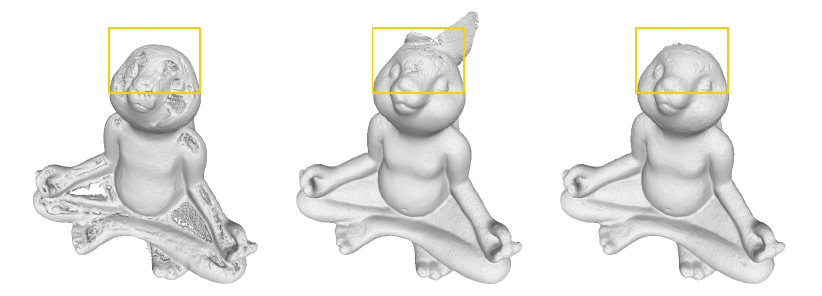}
        \put(5, -3){\small (a) w/o p-m}
        \put(40,-3){\small (b) w/o normal}
        \put(78,-3){\small (c) full}
   \end{overpic}
  \caption{\textbf{Ablation study}. Our full setting can achieve the best surface reconstruction results. 
  }
  \label{fig:ablation}
  \vspace{-8pt}
\end{figure}

%% file: table/ablation_v2.tex
\begin{table}
    \centering
    \caption{\textbf{Ablation studies of reconstruction quality on the DTU dataset.}
    We compare the average reconstruction performance on three settings: w/o normal priors, w/o patch-match guidance and our full setting. Our full setting achieves the best performance.
    }
    \begin{tabular}{ccccc}
    \hline\noalign{\smallskip}
    Model setting  & w/o normal & w/o p-m & full \\ 
    \noalign{\smallskip}
    \hline\noalign{\smallskip}
    CD & 0.53 & 0.93 & \textbf{0.52} \\
    \hline
    \end{tabular}

    \label{tab:ablation_study}
    \vspace{-8pt}
\end{table}

%% file: table/eval_mipnerf.tex
\begin{table}[t]
\vspace{-2pt}
\centering
\caption{Quantitative results on the Mip-NeRF360 \cite{barron2022mip} dataset. All baseline scores are taken directly from the respective papers, when available. 
}
\resizebox{0.98\columnwidth}{!}{
\begin{tabular}{@{}l|ccc|ccc}
\hline
 & \multicolumn{3}{c@{}|}{Outdoor Scene} & \multicolumn{3}{c@{}}{Indoor scene} \\ 
& PSNR~$\uparrow$ & SSIM~$\uparrow$ & LIPPS~$\downarrow$ & PSNR~$\uparrow$ & 
SSIM~$\uparrow$ & LIPPS~$\downarrow$ \\
\hline
NeRF & 21.46 & 0.458 & 0.515 & 26.84 &  0.790 & 0.370 \\
Deep Blending & 21.54 &0.524 & 0.364 & 26.40 & 0.844 & 0.261 \\
Instant NGP & 22.90 & 0.566 & 0.371 & 29.15 & 0.880 & 0.216 \\
MERF & 23.19 & 0.616 &  0.343 & 27.80 & 0.855 & 0.271 \\
BakedSDF & 22.47 & 0.585 &  0.349 & 27.06 & 0.836 & 0.258 \\
MipNeRF360 &   24.47 &  0.691 &  0.283 &   31.72 &  0.917 &  0.180 \\
\hline
\hline
Mobile-NeRF & 21.95 & 0.470 & 0.470 & - & - & - \\
SuGaR &  22.93 & 0.629 & 0.356 & 29.43 & 0.906 & 0.225 \\
3DGS &   \tbest24.64 &   \tbest0.731 &   0.234 &    \sbest30.41 &   \tbest0.920 &   0.189 \\
2DGS &  24.34 &   0.717 &  0.246  &  \tbest30.40 &  0.916 &  0.195  \\
GOF &  \sbest24.76 &   \sbest0.742 & \tbest0.225  & \best30.80 &  \sbest0.928 &  \sbest0.167  \\
PGSR &  24.45 &   0.730 &  \sbest0.224  &  \sbest30.41  & \best0.930  & \best 0.161  \\
Ours & \best 25.09 & \best0.753 &  \best0.212  &  30.05 & 0.920  & \tbest0.183 \\
\hline
\hline
\end{tabular}
}
\label{tab:mipnerf360}
\vspace{-2pt}
\end{table}

%% file: sec/5_conclusion.tex
\section{Conclusion}
In this paper,  we propose a novel approach for efficient and high-quality surface reconstruction while maintaining the capability of real-time novel view synthesis. We employ incorporating geometric guidance into the framework of 3D Gaussian Splatting. We leverage a patch-match-based Multi-View Stereo (MVS) technique
for geometric guidance at texture-rich image areas and normal prior guidance at texture-less image regions to improve reconstruction quality. Extensive experiments demonstrate the effectiveness of our design in surface reconstruction, novel view synthesis and training speed.